\documentclass[11pt]{article}

\usepackage{amsmath}
\usepackage{amssymb}
\usepackage{hyperref}
\usepackage{booktabs}
\usepackage{graphicx}
\usepackage{natbib}
\usepackage[margin=1in]{geometry}
\usepackage{listings}
\usepackage{xcolor}

\hypersetup{
  pdftitle={Authorization Propagation in Multi-Agent AI Systems: Identity Governance as Infrastructure},
  pdfauthor={Krti Tallam},
  pdfsubject={Workflow-scoped authorization architecture for multi-agent AI systems},
  pdfkeywords={agentic AI, authorization, delegation, access control, identity governance, multi-agent systems},
  colorlinks=true,
  linkcolor=blue,
  citecolor=blue,
  urlcolor=blue
}

\title{Authorization Propagation in Multi-Agent AI Systems:\\Identity Governance as Infrastructure}
\author{Krti Tallam\\Kamiwaza AI\\\texttt{krti@kamiwaza.ai}}
\date{May 2026}

\begin{document}
\maketitle

\begin{abstract}
The security discussion around agentic AI focuses heavily on prompt injection. This paper argues that multi-agent systems also create a distinct authorization problem: maintaining authorization invariants as non-human principals retrieve data, delegate tasks, and synthesize results across changing boundaries. We call this problem \emph{authorization propagation}. It is not reducible to prompt injection and is not fully addressed by classical access-control models such as RBAC, ABAC, or ReBAC. The paper formalizes authorization propagation as a workflow-level property, identifies three sub-problems (transitive delegation, aggregation inference, and temporal validity), and derives seven structural requirements for authorization architectures in multi-agent AI systems. Recent work on invocation-bound capability tokens \citep{prakash2026aip}, task-scoped authorization envelopes \citep{sharma2026pauth}, dependency-graph policy enforcement \citep{palumbo2026pcas}, and execution-count revocation \citep{parakhin2026bureaucracy} demonstrates that the field is converging on the problem, but not yet on a complete architecture. The central claim is that identity governance must be treated as infrastructure: evaluated continuously, enforced at every interaction boundary, and designed into the system before orchestration logic is allowed to scale. Preliminary implementation evidence from a production enterprise AI platform shows that ordinary system behavior, not only adversarial action, already produces the failures this model predicts.
\end{abstract}


\section{Introduction}

As AI systems evolve from single-model inference to multi-agent orchestration, the security discourse has concentrated heavily on prompt injection. This is understandable. Prompt injection is the mechanism by which an adversary subverts an agent's behavior by manipulating the content it processes. It is well-documented, broadly applicable, and --- as of this writing --- not fully solved at the model level.

The resulting industry posture treats prompt injection as the central, sometimes only, novel security problem in agentic AI. A common argument, articulated informally but widely held, runs approximately as follows: prompt injection is to agentic AI what SQL injection was to web applications. Fix it architecturally --- separate data from instructions, as parameterized queries separated data from SQL --- and the remaining security problems reduce to known primitives: authentication, authorization, network isolation, secrets management. Nothing novel.

This paper argues that the analogy is partially correct but materially incomplete. Prompt injection is the novel \emph{attack vector}. But there is a distinct novel \emph{architectural problem} that persists even under the assumption that prompt injection is fully solved: the problem of maintaining authorization invariants as non-human principals make chains of autonomous decisions.

We call this problem \emph{authorization propagation}.

Consider a concrete scenario. An orchestrating agent receives a user query. It decomposes the task and delegates sub-tasks to specialized agents. Agent A retrieves data from Dataset X. Agent B retrieves data from Dataset Y. Agent C synthesizes the results from A and B into a response for the user. Even with perfect prompt injection defense --- no agent can be tricked by the content it processes --- the system must answer:

\begin{itemize}
  \item Did Agent A have the right to access Dataset X?
  \item Did Agent B have the right to access Dataset Y?
  \item Does Agent C have the right to see the combined results?
  \item Does the user have the right to see the synthesized output?
  \item Does the \emph{combination} of X and Y reveal information that neither dataset alone would expose?
\end{itemize}

These are authorization questions, not prompt injection questions. They arise from the structure of the system, not from adversarial content. And they have no established equivalent of ``parameterized queries'' --- no single architectural primitive that resolves them.

Why this matters now is that enterprise agents no longer merely retrieve documents on behalf of a visible human operator. They increasingly mediate the entire evidentiary path: decomposition, retrieval, tool invocation, synthesis, and delivery. In that setting, authorization is no longer just a question of whether an individual access is permitted. It is a question of whether the full delegated workflow preserves the authority, scope, and boundary conditions that make the final result governable.

This paper formalizes authorization propagation, distinguishes it from prompt injection, and identifies what it demands of authorization architectures.

\section{Background and Related Work}

\subsection{Prompt Injection and Agent Security}

Prompt injection was first characterized as a distinct vulnerability class in 2022 \citep{gruskovnjak2023indirect} and has since been the subject of extensive research \citep{greshake2023compromising}. Google DeepMind's ``AI Agent Traps'' taxonomy \citep{deepmind2026agenttraps} identifies six categories of agent-directed attacks: content injection, semantic manipulation, cognitive state attacks, tool misuse induction, goal hijacking, and multi-agent collusion. These categories are useful but share a common structure: an adversary manipulates the content, context, or memory that an agent processes in order to alter its behavior.

The defense literature has correspondingly focused on content-level mitigations: input filtering, output validation, trusted/untrusted content separation, and instruction hierarchy enforcement.

What this literature does not address in depth is the authorization architecture that determines what data and actions are available to agents in the first place. The implicit assumption is that if agents can be made robust to adversarial content, the remaining security properties can be handled by conventional access control.

Recent empirical work challenges this assumption. \citet{anthropic2026claudecode} evaluate the Claude Code permission system and find an 81.0\% false negative rate on deliberately ambiguous authorization scenarios, with 36.8\% of state-changing actions bypassing the classifier entirely via file edits. \citet{debenedetti2026roleconfusion} demonstrate that prompt injection can be reframed as role confusion, with 60\% attack success on StrongREJECT via spoofed reasoning --- arguing that ``security is defined at the interface but authority is assigned in latent space.'' These results suggest that even well-resourced content-level defenses are insufficient, reinforcing the need for architectural authorization enforcement.

\subsection{Classical Access Control Models}

The access control literature provides foundational models that inform but do not fully address authorization propagation. \citet{bell1973secure} formalize mandatory access control with the $\star$-property (no write-down) and simple security property (no read-up), preventing information flow from high-classification to low-classification subjects. \citet{biba1977integrity} provides the dual integrity model: no read-down, no write-up. \citet{clark1987comparison} shift focus to well-formed transactions and separation of duty, requiring that data modifications pass through certified transformation procedures.

\citet{sandhu1996role} formalize role-based access control (RBAC), which assigns permissions to roles rather than directly to subjects. XACML \citep{xacml2013} extends attribute-based access control with obligation policies --- actions that must be performed upon access denial or grant.

These models address important properties: confidentiality (Bell-LaPadula), integrity (Biba), transaction correctness (Clark-Wilson), administrative scalability (RBAC), and policy expressiveness (XACML). Contemporary risk frameworks \citep{nist2023airmf, iso2023iec42001} acknowledge that AI systems introduce novel authorization challenges, but delegate the specifics to implementation standards that do not yet exist for multi-agent architectures. The classical models share assumptions that do not hold in multi-agent AI systems: human principals, static resources, synchronous access decisions, and well-defined trust boundaries. The authorization propagation problem arises precisely where these assumptions break down.

\subsection{Relationship-Based Access Control}

Relationship-based access control (ReBAC) originates from Google's Zanzibar system \citep{pang2019zanzibar}, which models authorization as a graph of typed relationships between subjects and objects. A subject's permissions are determined not by static role assignments (as in RBAC) or attribute predicates (as in ABAC), but by the existence and type of relationships in a tuple store. This enables fine-grained, context-sensitive authorization that can express ownership, delegation, group membership, and organizational hierarchy.

ReBAC has been implemented in several open systems (SpiceDB, OpenFGA, Authzed) and is well-suited to multi-tenant enterprise environments where access patterns are relational rather than role-based.

However, the Zanzibar model and its descendants were designed for human principals interacting with static resources through well-defined API surfaces. They do not natively address:

\begin{itemize}
  \item non-human principals that autonomously chain authorization decisions
  \item transitive delegation where one agent acts on behalf of another
  \item synthesized outputs where the result is derived from multiple authorized sources but the combination may not itself be authorized
  \item temporal validity where authorization state may change between the start and end of a multi-step agent workflow
\end{itemize}

\subsection{Emerging Agent Authorization Frameworks}

The period from late 2025 through early 2026 has seen a rapid emergence of authorization frameworks specifically designed for agentic AI systems. We survey the most architecturally significant.

\paragraph{Invocation-bound capability tokens.} \citet{prakash2026aip} proposes Invocation-Bound Capability Tokens (IBCTs) that fuse identity, attenuated authorization, and provenance binding into an append-only token chain. Two wire formats are specified: compact JWT for single-hop and Biscuit tokens with Datalog policies for multi-hop delegation. Reference implementations in Python and Rust demonstrate 0.049ms verification latency and 100\% adversarial rejection across 600 attack attempts. IBCTs address R2 (explicit, bounded, auditable delegation) and partially address R5 (self-contained authorization traces), but do not address aggregation inference (R4).

\paragraph{Task-scoped authorization envelopes.} \citet{sharma2026pauth} propose PAuth, which challenges OAuth's operator-scoped authorization model. PAuth introduces ``NL slices'' --- symbolic specifications of expected tool calls per service derived from natural language task descriptions --- and ``envelopes'' that bind operand values to symbolic provenance. The system achieves 100\% success on benign tasks and 100\% warning rate on injected attacks. The envelope concept is architecturally adjacent to the execution envelope pattern described in the companion fail-and-report paper, and the NL-slice concept provides a mechanism for deriving per-task authorization scope.

\paragraph{Capability coherence via execution counting.} \citet{parakhin2026bureaucracy} provides formal proof that TTL-based token revocation fails at agent execution speeds. Mapping CPU cache coherence protocols (MESI) to agent authorization revocation, the paper demonstrates that execution-count-based Release Consistency bounds unauthorized operations at $D_{rcc} \leq n$ \emph{independent of agent velocity}, versus $O(v \cdot TTL)$ scaling for time-based approaches --- a 120$\times$ reduction in unauthorized operations. This directly addresses the temporal validity sub-problem (Section~\ref{sec:temporal}) by demonstrating that time-based revocation is architecturally wrong for autonomous agents.

\paragraph{Dependency-graph policy enforcement.} \citet{palumbo2026pcas} propose PCAS, which models agent state as a dependency graph capturing causal relationships among tool calls, results, and messages. Datalog-derived policies are enforced by a reference monitor. The system improves policy compliance from 48\% to 93\% across frontier models with zero policy violations under deterministic enforcement independent of model reasoning. The dependency graph approach provides a concrete mechanism for R3 (authorization at every retrieval boundary) and R5 (workflow-scoped traces).

\paragraph{Standards-track proposals.} The \citet{openid2025agentic} has published a consensus whitepaper on agentic identity, and \citet{benameur2025oidca} propose OIDC-A (OpenID Connect for Agents 1.0), extending OIDC with agent identity, delegation chain validation, attestation verification, and capability-based authorization. Together with the IETF AIP track \citep{prakash2026aip}, these represent the two major standards efforts.

\subsection{Formal Methods for Agent Security}

Several recent works provide formal foundations relevant to authorization propagation.

\citet{chen2026aith} proposes AITH, a post-quantum continuous delegation protocol for AI agents. Its six-check Boundary Engine implements push-based revocation within one second. All five security theorems are machine-verified via the Tamarin Prover under the Dolev-Yao model. Notably, the protocol achieves 79.5\% autonomous execution with 6.1\% human escalation and 14.4\% blocked --- the escalation rate representing an implementation of the fail-and-report pattern where the system halts and reports rather than proceeding under uncertain authorization.

\citet{garby2026llmbda} present the LLMbda Calculus, a lambda calculus enriched with dynamic information-flow control and LLM invocation primitives. Their termination-insensitive noninterference theorem establishes integrity and confidentiality guarantees for agentic programming. Noninterference --- the property that low-integrity inputs cannot influence high-integrity actions --- is the formal analogue of the authorization propagation guarantee: information flow through the agent chain preserves authorization boundaries.

\citet{song2026formalizing} (UC Berkeley / ETH Zurich) propose four contextual security properties for LLM agents: task alignment, action alignment, source authorization, and data isolation. Source authorization maps directly to authorization propagation; data isolation maps to the anti-aggregation-inference requirement. The formalization provides an independent validation of the problem decomposition presented in this paper.

\subsection{Empirical Evidence of Authorization Failures}

The authorization propagation problem is not merely theoretical. Recent empirical work demonstrates that it produces real, exploitable failures.

\paragraph{Semantic intent fragmentation.} The SIF attack \citep{ahad2026semantic}, accepted at the AAAI 2026 Summer Symposium, decomposes a legitimate request into individually benign subtasks that jointly violate security policy through four mechanisms: bulk scope escalation, silent data exfiltration, embedded trigger deployment, and quasi-identifier aggregation. A 71\% success rate across 14 enterprise scenarios provides direct empirical evidence that aggregation inference (Section~\ref{sec:aggregation}) is a real, exploitable attack vector, not a theoretical concern.

\paragraph{Agentic deanonymization.} \citet{li2026agentic} demonstrates that LLM agents with web search can link anonymized interview data to specific individuals by decomposing re-identification into individually benign sub-tasks that bypass safeguards. This is aggregation inference in practice: each individual data access is authorized, but their composition violates privacy policy.

\paragraph{Production deployment evidence.} \citet{maiti2026healthcare} reports on 9 autonomous AI agents deployed in a healthcare production environment for 90 days under zero-trust constraints. The deployment uses gVisor isolation, credential proxy sidecars, egress allowlisting, and structured metadata envelopes. Four HIGH severity findings from automated security testing were identified, providing rare production-scale evidence that authorization architecture matters in practice. The credential proxy sidecar pattern implements authorization propagation at the container layer.

\paragraph{MCP threat landscape.} \citet{mcpshield2026} identifies 7 threat categories, 23 attack vectors, and 4 attack surfaces across 177{,}000+ MCP tools, finding that no single defense covers more than 34\% of threats. An integrated architecture achieves 91\%, providing quantitative evidence for the defense-in-depth argument and against the notion that any single authorization mechanism is sufficient.

\subsection{The Companion Governance Framework}

The companion paper, \emph{Machine-Speed Accountability} \citep{tallam2026machine}, provides a governance operating model for oversight-heavy AI environments. It defines four constraints --- traceability, auditability, controllability, and recovery --- and argues that accountability must be treated as a system constraint rather than a reporting layer. Its control layer identifies the seams where controls must exist: between data and retrieval, model and context, model and tools, tools and systems of record, and human operators and delegated agents.

This paper provides the technical depth for the last of those seams --- the boundary between human operators and delegated agents, and between agents themselves --- which is where authorization propagation occurs.

\section{A Taxonomy: Attack Vectors and Architectural Problems}

The current AI security discourse conflates two categories of challenge that have different structures, different solutions, and different implications for system design.

\subsection{Attack Vectors}

Attack vectors are mechanisms by which an adversary subverts system behavior. In agentic AI, the primary attack vectors are:

\begin{itemize}
  \item \textbf{Content injection}: adversarial instructions embedded in data that an agent processes (e.g., hidden text in HTML, manipulated documents)
  \item \textbf{Semantic manipulation}: authoritative-sounding content that causes an agent to override its instructions
  \item \textbf{Cognitive state attacks}: poisoning an agent's memory, context, or learned state
  \item \textbf{Tool misuse induction}: manipulating an agent into invoking tools in unintended ways
\end{itemize}

These share a common property: they require an adversary to introduce malicious content into the agent's processing path. Defense is a content-integrity problem.

\subsection{Architectural Problems}

Architectural problems are structural challenges that arise from the design of the system itself, independent of adversarial action. In multi-agent AI, the primary architectural problems are:

\begin{itemize}
  \item \textbf{Authorization propagation}: maintaining access control invariants as agents delegate tasks, retrieve data, and synthesize results across multiple authorization boundaries
  \item \textbf{Transitive delegation}: determining what authority an agent inherits when acting on behalf of another agent or a human principal
  \item \textbf{Aggregation inference}: determining whether a synthesized result derived from individually authorized data sources is itself authorized for the requesting principal
  \item \textbf{Temporal validity}: ensuring that authorization decisions remain valid across the duration of a multi-step autonomous workflow
\end{itemize}

A recent systematization of knowledge \citep{ruohonen2025sok} introduces the Privilege Escalation Distance metric for quantifying authorization gaps across agentic attack surfaces. These architectural problems do not require an adversary. They arise from normal system operation. A misconfigured agent chain, a stale relationship tuple, a workroom membership change that propagates too slowly --- any of these can produce unauthorized data exposure without a traditional breach.

\subsection{Why the Distinction Matters}

Solving prompt injection does not solve authorization propagation. Even in a system with perfect content integrity --- where no agent can be tricked by anything it reads --- the authorization questions remain:

\begin{itemize}
  \item Which data sources is each agent permitted to access?
  \item Does delegation from Agent A to Agent B carry A's authority, B's authority, or neither?
  \item When Agent C synthesizes results from multiple sources, does the user have authorization for the synthesis?
  \item If a relationship tuple is revoked mid-workflow, what happens to results already derived from data accessed under that tuple?
\end{itemize}

Conversely, solving authorization propagation does not solve prompt injection. A system with perfect authorization architecture can still be subverted by an adversary who manipulates agent behavior through content injection.

The two problems are complementary, not reducible. Treating them as a single problem --- or treating one as a subset of the other --- leads to architectures that are robust against attacks but structurally unsound, or structurally sound but vulnerable to adversarial content.

\citet{bhattarai2026deterministic} independently arrive at a compatible conclusion, identifying the ``Lethal Trifecta'' --- untrusted inputs, privileged data access, and external action capability --- and arguing that alignment is insufficient; architectural mediation is required. Their finite action calculus formalizes monotonic permissions, complementing the structural requirements presented in Section~\ref{sec:requirements}.

\section{Identity Governance as Infrastructure}

\subsection{The Convergence Thesis}

We observe that the authorization challenges in multi-agent AI systems converge from two independent directions:

\paragraph{Internal drift.} Multi-agent systems accumulate authorization inconsistencies through normal operation. Ontologies evolve. Relationship tuples go stale. Workroom or team membership changes. Agent configurations are updated without corresponding authorization reviews. The system drifts into a state where its actual data access patterns no longer match its intended authorization posture. No attacker is required.

\paragraph{External attack.} The DeepMind Agent Traps taxonomy \citep{deepmind2026agenttraps} demonstrates that the same system layers that drift internally --- context, memory, tool interfaces, data retrieval paths --- are also the attack surface for external adversaries. An attacker who can manipulate an agent's context can alter what data it retrieves. An attacker who can poison an agent's memory can alter what it considers authorized.

Both vectors produce the same outcome: unauthorized data exposure without a traditional perimeter breach. The convergence of internal drift and external attack on the same system layers is what elevates identity governance from a policy concern to an infrastructure requirement.

\subsection{What ``Infrastructure'' Means Here}

To say identity governance is infrastructure is to make a specific architectural claim: authorization state must be evaluated continuously and enforced at every interaction boundary in the system, not checked periodically or applied as a middleware layer.

The analogy is to encryption in transit. Early web architectures treated TLS as optional, applied selectively to ``sensitive'' endpoints. Modern architectures treat it as infrastructure --- always on, not a decision point. The reason is that the cost of deciding ``does this particular request need encryption?'' exceeded the cost of encrypting everything, and the consequences of deciding wrong were unacceptable.

Identity governance in multi-agent AI is at an analogous inflection. The cost of deciding ``does this particular agent interaction need authorization checking?'' exceeds the cost of checking everything, and the consequences of deciding wrong --- data leakage through an unchecked agent hop --- are unacceptable in oversight-heavy environments.

This means authorization evaluation cannot be a gateway that agents pass through once. It must be a property of every data retrieval, every delegation, and every result synthesis in the system.

\section{Authorization Propagation: The Formal Problem}

\subsection{System Model}

Consider a multi-agent system with the following components:

\begin{itemize}
  \item A set of \textbf{human principals} $H = \{h_1, h_2, \ldots, h_m\}$
  \item A set of \textbf{agent principals} $A = \{a_1, a_2, \ldots, a_n\}$
  \item A set of \textbf{data resources} $D = \{d_1, d_2, \ldots, d_k\}$
  \item An \textbf{authorization function} $\mathrm{auth}(s, r, t) \rightarrow \{\mathrm{allow}, \mathrm{deny}\}$ that determines whether subject $s$ may access resource $r$ at time $t$
  \item A \textbf{delegation function} $\mathrm{delegate}(s_1, s_2, t) \rightarrow P$ that determines what permissions principal $s_1$ transfers to principal $s_2$ at time $t$
\end{itemize}

A \textbf{workflow execution} is a directed acyclic graph $W = (V, E)$ where each vertex $v \in V$ is an agent action (retrieve, transform, synthesize, or return) and each edge $(v_i, v_j)$ represents data flow from action $v_i$ to action $v_j$.

\subsection{The Three Sub-Problems}

\subsubsection{Transitive Delegation}
\label{sec:delegation}

When a human principal $h$ initiates a workflow that involves agents $a_1, a_2, \ldots, a_n$, the system must determine the effective authority at each agent. The naive approach --- each agent inherits the initiating human's full authority --- violates least privilege. The conservative approach --- each agent operates only under its own service identity --- may be insufficient to complete the workflow.

The problem is to define a delegation model where:

\begin{itemize}
  \item Each agent's effective authority is bounded and auditable
  \item Authority does not accumulate through delegation chains (no privilege escalation)
  \item The initiating principal's authorization is necessary but not sufficient for downstream access
  \item Delegation can be revoked at any point in the chain
\end{itemize}

No widely deployed authorization system provides a complete solution to this in the context of autonomous agent-to-agent delegation. The confused deputy problem --- where an agent is tricked into misusing its authority on behalf of an unauthorized principal --- is a classical formulation of this failure mode, and has been identified in multi-agent LLM systems specifically \citep{seagent2026}. However, emerging work on invocation-bound capability tokens \citep{prakash2026aip} and post-quantum delegation chains \citep{chen2026aith} demonstrates that the token-level mechanisms are feasible. \citet{prakash2026aip} achieves 0.049ms per-boundary verification with append-only attenuation, and \citet{chen2026aith} provides Tamarin-verified revocation within one second. What remains is integration into a workflow-scoped authorization architecture.

\subsubsection{Aggregation Inference}
\label{sec:aggregation}

Even when each individual data access in a workflow is independently authorized, the synthesis of results may produce information that no individual access would expose. This is the aggregation inference problem.

Formally: given authorized accesses to resources $d_1, d_2, \ldots, d_j$ and a synthesis function
\[
f(d_1, d_2, \ldots, d_j) \rightarrow r,
\]
is the result $r$ authorized for the requesting principal?

This problem is not new --- it appears in statistical database security and classified information handling (the ``mosaic effect''). But it takes on new dimensions in agentic AI because:

\begin{itemize}
  \item The synthesis function $f$ is a neural network whose behavior is not formally specified
  \item The set of accessed resources may not be known in advance (agents may discover and access resources dynamically)
  \item The requesting principal may not know what resources contributed to the result
\end{itemize}

Recent empirical work confirms that aggregation inference is not merely theoretical. The Semantic Intent Fragmentation attack \citep{ahad2026semantic} achieves a 71\% success rate across 14 enterprise scenarios by decomposing legitimate requests into individually benign subtasks that jointly violate security policy. \citet{li2026agentic} demonstrates that LLM agents can link anonymized interview data to specific individuals through individually authorized web searches, establishing aggregation inference as a practical privacy attack. These results elevate aggregation inference from a known theoretical concern to a demonstrated exploitation technique.

The aggregation inference problem is genuinely unsolved in the general case. What can be addressed architecturally is (a) making the set of contributing resources inspectable after the fact, (b) defining policy boundaries that limit which resource combinations are permitted, and (c) tracking causal dependencies through the synthesis graph --- an approach demonstrated by the PCAS dependency-graph monitor \citep{palumbo2026pcas}, which improves policy compliance from 48\% to 93\%.

\subsubsection{Temporal Validity}
\label{sec:temporal}

Multi-agent workflows execute over time. A workflow initiated at time $t_0$ may not complete until $t_n$. Authorization state may change during execution: a relationship tuple may be revoked, a principal's role may change, a data resource may be reclassified.

The question is: at what time is authorization evaluated?

\begin{itemize}
  \item \textbf{Initiation-time}: authorization is checked when the workflow starts and assumed valid throughout. This is simple but may permit access to data that was subsequently restricted.
  \item \textbf{Access-time}: authorization is checked at each data retrieval. This is correct but may cause workflows to fail mid-execution, requiring partial rollback.
  \item \textbf{Completion-time}: authorization is checked before results are returned to the initiating principal. This prevents unauthorized delivery but may waste compute on workflows whose results cannot be disclosed.
\end{itemize}

Each policy has different failure modes and different implications for the recovery constraint defined in \citep{tallam2026machine}. The choice is a governance decision, not a purely technical one.

\citet{parakhin2026bureaucracy} demonstrates quantitatively why time-based revocation (the JWT TTL model) fails at agent execution speeds. Mapping CPU cache coherence protocols to authorization revocation, the analysis shows that TTL-based approaches produce unauthorized operations scaling as $O(v \cdot TTL)$ where $v$ is agent velocity, while execution-count-based Release Consistency bounds violations at $D_{rcc} \leq n$ independent of velocity --- a 120$\times$ reduction. This argues strongly against initiation-time evaluation for any system where agents operate at machine speed, and suggests that access-time evaluation with execution-count-based validity checks is the architecturally sound default.

\subsection{A Worked Authorization Chain}

Consider a due-diligence workflow with four principals: a human analyst, an orchestrating agent, a retrieval agent, and a synthesis agent.

\begin{enumerate}
  \item The analyst asks for a summary of liabilities that could materially change valuation.
  \item The orchestrator decomposes the request into financial retrieval, legal-annex retrieval, and review-committee memo retrieval.
  \item The retrieval agent has access to the deal room, but not to the restricted review-committee workspace that contains a contingent-liability memo.
  \item A synthesis agent combines the retrieved materials and returns a valuation-risk summary to the analyst.
\end{enumerate}

Every stage raises a different authorization question. The orchestrator may be allowed to delegate the legal-annex retrieval but not the restricted memo retrieval. The retrieval agent may be allowed to access each visible corpus independently while still lacking authority to create a cross-boundary synthesis context. The synthesis agent may be permitted to read both outputs only if the delegated scope remains intact and no boundary has been crossed under stale authority. The analyst may be authorized to receive a partial result, but not a result presented as complete when a material source was excluded.

This is the core reason authorization propagation is not reducible to per-request authentication or per-access policy checks. The governable object is the chain itself: who initiated it, which authority was delegated, what data crossed which boundaries, what was combined, and under what validity regime that combination remained authorized.

\section{Why Existing Authorization Models Are Insufficient}

\subsection{RBAC (Role-Based Access Control)}

RBAC \citep{sandhu1996role} assigns permissions to roles, and roles to users. It does not model relationships between resources, does not support transitive delegation natively, and cannot express ``this agent may access this dataset only when acting on behalf of a specific user within a specific workflow.'' The granularity is wrong for multi-agent authorization.

\subsection{ABAC (Attribute-Based Access Control)}

ABAC evaluates access decisions based on attributes of the subject, resource, action, and environment. It is more expressive than RBAC but treats each access decision independently. It does not model the \emph{chain} of accesses that constitutes a multi-agent workflow, and cannot express constraints on the aggregation of individually authorized accesses. XACML obligation policies \citep{xacml2013} can trigger actions upon access denial, but obligations are evaluated per-access, not per-workflow --- they cannot enforce constraints that span the full delegation and synthesis graph.

\subsection{ReBAC (Relationship-Based Access Control)}

ReBAC is the closest existing model to what multi-agent authorization requires. Its graph-based relationship model can express delegation, group membership, and hierarchical access. However, as currently specified and implemented:

\begin{itemize}
  \item Relationships are between human principals and static resources. Non-human agent principals are not first-class entities in most ReBAC schemas.
  \item Delegation is modeled as a static relationship, not as a dynamic, workflow-scoped authority transfer.
  \item There is no native concept of workflow-scoped authorization that binds a set of accesses together and constrains their aggregation.
  \item Temporal validity is handled through tuple creation and deletion, not through workflow-aware authorization epochs.
\end{itemize}

ReBAC provides the right \emph{primitives} --- relationship tuples, graph-based permission evaluation, fine-grained resource binding --- but requires extension to address authorization propagation in multi-agent contexts.

\subsection{Emerging Models and Remaining Gaps}

The frameworks surveyed in Section 2.4 address fragments of the problem space. IBCTs \citep{prakash2026aip} solve per-boundary delegation with attenuation. PAuth \citep{sharma2026pauth} solves task-scoped authorization derivation. PCAS \citep{palumbo2026pcas} solves causal dependency tracking. AITH \citep{chen2026aith} solves post-quantum revocation. No single framework satisfies all seven structural requirements (Section~\ref{sec:requirements}).

The gap is integration. A sufficient authorization architecture for multi-agent AI must compose: (a) append-only delegated authority (IBCTs), (b) task-scoped authorization derivation (PAuth/NL-slices), (c) causal dependency tracking for aggregation (PCAS), (d) execution-count-based temporal validity \citep{parakhin2026bureaucracy}, and (e) workflow-scoped traces (R5). Whether these mechanisms can be composed without introducing new failure modes is an open architectural question.

\section{Instantiating the Accountability Constraints}

The companion paper \citep{tallam2026machine} defines four constraints for oversight-heavy AI: traceability, auditability, controllability, and recovery. Authorization propagation provides a specific, technically demanding instantiation of each.

\subsection{Traceability}

For authorization propagation, traceability requires that the system can reconstruct:

\begin{itemize}
  \item The initiating principal and their authorization state at workflow initiation
  \item The delegation chain: which agent delegated to which, with what authority transfer
  \item The data access chain: which agent accessed which resource, under what authorization, at what time
  \item The synthesis chain: which results were combined, by which agent, to produce the final output
\end{itemize}

This is more demanding than logging individual API calls. It requires a \emph{workflow-scoped authorization trace} that links the initiating authority to every downstream access and synthesis.

\subsection{Auditability}

An independent reviewer must be able to inspect the authorization trace and determine:

\begin{itemize}
  \item Whether each individual access was authorized at the time it occurred
  \item Whether the delegation chain was valid (no privilege escalation, no stale authority)
  \item Whether the aggregation of accesses was within policy bounds
  \item Whether the final result was authorized for delivery to the requesting principal
\end{itemize}

This requires that the authorization trace is self-contained --- the reviewer should not need to replay the workflow or query live authorization state to reach a conclusion. Tamper-evident receipts \citep{errico2026aarm} provide one mechanism for ensuring trace integrity.

\subsection{Controllability}

The system must be able to:

\begin{itemize}
  \item Deny a downstream agent access even if the upstream agent was authorized (least privilege at each hop, not inherited privilege)
  \item Pause or terminate a workflow when an authorization boundary is crossed mid-execution
  \item Degrade gracefully: return partial results from authorized sources rather than failing entirely or returning unauthorized results
\end{itemize}

\subsection{Recovery}

When an authorization violation is discovered retroactively:

\begin{itemize}
  \item The system must be able to identify all downstream results that were derived from the unauthorized access
  \item Results that were delivered to principals must be flaggable for review
  \item Authorization tuples must be correctable, and the correction must propagate to prevent recurrence
\end{itemize}

The difficulty of recovery scales with the depth and breadth of the agent chain. A single unauthorized access at an early hop may taint every downstream result. This is why authorization propagation must be treated as a design-time architectural concern, not a runtime patching exercise.

\section{Structural Requirements}
\label{sec:requirements}

We do not propose a specific authorization architecture. We identify the structural requirements that any sufficient architecture must satisfy.

\textbf{R1. Agent principals must be first-class authorization subjects.} Agents must have scoped identities with explicit, bounded permissions. Service accounts with broad access are insufficient.

\textbf{R2. Delegation must be explicit, bounded, and auditable.} When an agent delegates to another agent, the authority transfer must be recorded, scoped to the workflow, and subject to policy constraints. Implicit authority inheritance through shared credentials or ambient permissions violates least privilege.

\textbf{R3. Authorization must be evaluated at every data retrieval boundary.} Not once at workflow initiation, but at every point where an agent accesses a data resource. The evaluation must be against the agent's effective authority in the context of the current workflow, not against a static permission set.

\textbf{R4. Aggregation policies must be expressible and enforceable.} The system must be able to define constraints on which resource combinations are permitted for a given principal, and enforce those constraints before synthesized results are returned.

\textbf{R5. Authorization traces must be workflow-scoped and self-contained.} The complete authorization history of a workflow --- from initiation through every delegation, access, and synthesis --- must be capturable as a single, inspectable artifact.

\textbf{R6. Temporal validity must be a policy decision, not an implementation default.} The system must support configurable authorization evaluation policies (initiation-time, access-time, completion-time) with explicit tradeoff documentation.

\textbf{R7. Recovery must be traceable through the synthesis graph.} When an authorization violation is discovered, the system must be able to identify all results derived from the violated access, across all downstream agents and synthesis steps.

\section{Discussion}

\subsection{The SQL Injection Analogy, Revisited}

The informal argument that prompt injection is to agentic AI what SQL injection was to web applications has real merit. SQL injection was solved architecturally by separating data from instructions (parameterized queries), not by trying to sanitize inputs. Prompt injection likely requires an analogous architectural separation --- perhaps explicit trust boundary tokens in the model's input processing, or formal separation of instruction and data channels.

But SQL injection was not the only security problem in web applications. Authentication, authorization, session management, and access control remained hard problems even after parameterized queries were universally adopted. The existence of an architectural fix for the injection vector did not eliminate the need for authorization architecture.

The same applies here. An architectural fix for prompt injection --- if achieved --- would eliminate one important attack vector. It would not address authorization propagation, because authorization propagation is not caused by adversarial content. It is caused by the structure of multi-agent systems themselves.

\subsection{The Governance Implication}

The companion paper \citep{tallam2026machine} argues that accountability is a system constraint, not a reporting layer. Authorization propagation provides the clearest illustration of why.

In a single-agent system, authorization is a boundary check. The user is authorized (or not) to access a resource. In a multi-agent system, authorization is a \emph{flow property} --- it must hold at every point in a directed graph of delegations, accesses, and syntheses. Boundary checks are necessary but insufficient. The system must reason about authorization as a property of the workflow, not just of individual accesses.

This is why identity governance is infrastructure. It cannot be added after the agent orchestration architecture is designed. It must be a first-class design constraint, co-equal with the functional requirements of the system. Attempting to retrofit authorization propagation onto an existing multi-agent architecture is analogous to attempting to retrofit encryption into a protocol designed without it --- technically possible, but the resulting system is fragile, auditable only with difficulty, and prone to subtle failures.

\subsection{What Current Platforms Need}

For current agentic platforms, the paper's thesis cashes out into a small set of concrete requirements.

\textbf{First-class non-human identity and delegation artifacts.} Agent identity cannot remain an ambient service-account property. Systems need explicit delegation artifacts that preserve requester identity, delegated scope, and attenuation across hops \citep{prakash2026aip,tallam2026fromcantowould}.

\textbf{Stable workflow receipts and admission surfaces.} If a later reviewer cannot reconstruct which execution path was admitted, narrowed, redirected, or denied, then authorization propagation collapses into guesswork. Shared admission or receipt-bearing surfaces are therefore not mere observability conveniences; they are part of the authorization substrate \citep{tallam2026executionenvelope,errico2026aarm}.

\textbf{A measurable surface for unsafe completeness.} One reason authorization failures are under-specified is that many evaluation stacks do not measure them directly. Benchmark work on authorization-limited evidence provides an empirical surface for whether systems silently overclaim completeness, conservatively underclaim, or surface insufficiency in an operationally usable way \citep{tallam2026partialbench,tallam2026failandreport}.

\textbf{A disciplined degraded path.} Not every authorization shortfall should become a hard stop, and not every system can tolerate one. But if a platform chooses degradation, it needs a principled degraded path with explicit accountability semantics rather than ambient narrowing and implicit trust transfer \citep{tallam2026failandreport}.

\subsection{What Remains Unsolved}

This paper identifies the problem and the structural requirements. It does not claim to solve the aggregation inference problem in the general case --- this is an open research question with roots in statistical disclosure control and classified information handling that predates AI systems. Nor does it claim that the temporal validity problem has a single correct policy --- the right choice depends on the risk tolerance, operational requirements, and regulatory context of the deploying organization.

The rapid emergence of frameworks surveyed in Section 2.4 demonstrates that the community is converging on the problem. IBCTs \citep{prakash2026aip}, PAuth \citep{sharma2026pauth}, PCAS \citep{palumbo2026pcas}, and AITH \citep{chen2026aith} each solve fragments. But the gap between fragment solutions and a unified authorization propagation architecture is substantial. The MCPSHIELD finding \citep{mcpshield2026} --- that no single defense covers more than 34\% of threats --- applies by analogy: no single authorization mechanism addresses all seven structural requirements.

The LLMbda Calculus \citep{garby2026llmbda} offers the most promising formal foundation, with its termination-insensitive noninterference theorem providing the theoretical basis for provable authorization flow guarantees. Whether the formal model can be operationalized into a practical authorization architecture at production scale remains an open question.

What this paper does claim is that these problems must be \emph{named}, \emph{distinguished from prompt injection}, and \emph{addressed architecturally} rather than assumed away.

\subsection{Limits and Non-Goals}

This paper does not present a complete deployment architecture, and its claims are narrower than they may first appear.

\begin{itemize}
  \item It does \textbf{not} solve aggregation inference in the general case. It identifies why it becomes unavoidable in multi-agent systems and what a sufficient architecture must make inspectable.
  \item It does \textbf{not} replace prompt-injection defenses. A system may solve authorization propagation badly, well, or not at all independently of its content-integrity posture.
  \item It does \textbf{not} propose a final policy language. The paper is about architectural requirements, not about a complete syntax or theorem of authorization composition.
  \item It does \textbf{not} assume agents are stable units. In practice, mutable orchestration, runtime rebinding, and self-modifying agents complicate identity continuity further \citep{tallam2026layeredmutability}. That complication is acknowledged here, not solved here.
\end{itemize}

\subsection{Preliminary Evidence from Implementation}

While this paper does not propose a specific authorization architecture, we have observed the structural requirements articulated in Section~\ref{sec:requirements} being violated in a production enterprise AI platform during a stabilization cycle. These violations were not the result of adversarial action. They arose from normal system operation --- configuration, integration, and error-handling decisions made by competent engineers under ordinary conditions. The bugs were identified and fixed through standard development process. We report them here not as failures of engineering quality but as empirical evidence that the authorization propagation problem produces real, observable consequences in real systems.

\subsubsection{Silent Scope Widening Under Session Failure}

A reverse-proxy gateway delegated authentication to an external auth service via a ForwardAuth mechanism. When session binding to a specific collaboration workspace failed, the middleware silently fell back from workspace-scoped authentication to a global authentication context.

This is a temporal validity violation (Section~\ref{sec:temporal}). The authorization scope established at session initiation became invalid when the binding failed, but the system continued operating under a wider scope. The user was authenticated but not scoped. Subsequent requests accessed resources outside the user's intended authorization boundary.

The fix required fail-closed behavior at the middleware layer: if workspace session binding fails, the request fails. The system must not continue under a wider scope.

\textbf{Structural requirement violated:} R3 (authorization at every data retrieval boundary) and R6 (temporal validity as policy decision). The middleware defaulted to a permissive fallback rather than treating scope loss as an authorization failure.

\subsubsection{Delegation Binding Reported as Successful When It Was Not}

A collaboration workspace entry API returned HTTP 200 (success) even when the session binding that should establish workspace-scoped authorization context failed to complete. Subsequent user actions proceeded without workspace-scoped authorization, operating in an unscoped context while the user believed workspace isolation was active.

This is a transitive delegation failure (Section~\ref{sec:delegation}). The user delegates authority to the workspace context. If the delegation binding fails but the system reports success, every subsequent action carries authority that was never properly established. The delegation is nominal but not actual.

The fix required treating session binding as a precondition, not an optimization. If the binding fails, the entry fails.

\textbf{Structural requirement violated:} R2 (delegation must be explicit, bounded, and auditable). The system reported successful delegation when no delegation had occurred. The audit record showed a delegation that did not exist in practice.

\subsubsection{Infrastructure Identity Gap}

Extension components (agent containers that operate within the platform) were not built through an attested build pipeline. The platform's application-layer identity mechanism (trusted requester context) verified the caller's identity at the service layer, but the infrastructure layer --- the container image itself --- was not attested. An extension container could be built from an unverified image while still presenting valid application-layer credentials.

This is an identity integrity gap. R2 requires that delegation is auditable and bounded; a delegation chain that includes an unattested container has a gap in the identity proof. The application layer says ``this request comes from Extension X.'' But the infrastructure layer cannot attest that Extension X is the genuine, unmodified extension.

The fix adopted a dual-build pattern using attested base images, closing the identity chain from application layer to infrastructure layer.

\textbf{Structural requirement implicated:} R2 (delegation auditable) and R5 (authorization traces self-contained). The trace was incomplete because it could not prove the infrastructure-layer identity of the delegating component.

\subsubsection{Observations}

Three observations emerge from these cases:

First, \textbf{all three violations arose from normal system operation}, not from adversarial action. This confirms the paper's argument that authorization propagation is an architectural problem, not an attack-vector problem. No adversary was involved. The system drifted into authorization-violating states through reasonable engineering decisions (graceful degradation under failure, optimistic success reporting, independent build pipelines).

Second, \textbf{all three violations produced the same class of outcome}: a principal operated with authorization broader than intended, without visibility into the discrepancy. This is the unauthorized-data-exposure-without-breach scenario described in Section~4.1. It supports the convergence thesis: whether the cause is internal drift or external attack, the failure presents identically at the authorization layer.

Third, \textbf{the fixes all converged on the same pattern}: fail closed at the boundary, treat authorization preconditions as preconditions rather than optimizations, and close identity gaps in the enforcement chain. This is consistent with the paper's argument that authorization propagation demands infrastructure-grade enforcement, not policy-layer mitigation.

We note that three cases from a single system do not constitute proof of the formal model. But they demonstrate that the structural requirements (R1--R7) identify real failure modes in production systems, and that violations of these requirements produce the authorization-boundary failures the model predicts.

\section{Future Work}

Several next steps would make the paper's argument more operational and more falsifiable.

\textbf{Interoperable delegation carriers.} The field needs delegation artifacts that survive across model, tool, service, and container boundaries without collapsing back into ambient credentials. Existing token and envelope proposals point in that direction, but interoperability remains open \citep{prakash2026aip,sharma2026pauth,tallam2026executionenvelope}.

\textbf{Authorization-aware evaluation.} Current agent benchmarks over-index on capability and under-specify governance failures. A more complete evaluation program should measure silent narrowing, unsafe completeness, delegation drift, and stale-authority behavior under realistic task decompositions \citep{tallam2026partialbench}.

\textbf{Mutable-agent identity.} As agents become more stateful, updatable, and self-modifying, the question ``what is the same principal over time?'' becomes materially harder. Authorization propagation and identity drift should be studied together rather than treated as separate concerns \citep{tallam2026layeredmutability}.

\textbf{Delegated intent models.} Multi-agent systems increasingly need to distinguish what an agent technically can do from what it is conditionally authorized to do on behalf of a specific requester. That suggests deeper integration between propagation models and identity-conditioned delegation work \citep{tallam2026fromcantowould}.

\section{Conclusion}

The security discourse around multi-agent AI systems has concentrated on prompt injection as the singular novel challenge. This paper argues that authorization propagation is a distinct, complementary problem that persists even under the assumption of perfect prompt injection defense. It arises not from adversarial content but from the structure of multi-agent systems: non-human principals making chains of autonomous decisions involving data retrieval, delegation, and synthesis across authorization boundaries.

We have presented a taxonomy distinguishing attack vectors from architectural problems, formalized authorization propagation and its three sub-problems (transitive delegation, aggregation inference, and temporal validity), shown where existing authorization models are insufficient, and identified seven structural requirements for authorization architectures in multi-agent AI systems. Preliminary implementation evidence from a production platform confirms that these requirements identify real failure modes, and that violations produce the authorization-boundary failures the model predicts.

The rapid emergence of fragment solutions demonstrates that the community is converging on the problem. Recent examples include invocation-bound capability tokens \citep{prakash2026aip}, task-scoped authorization envelopes \citep{sharma2026pauth}, execution-count revocation \citep{parakhin2026bureaucracy}, dependency-graph enforcement \citep{palumbo2026pcas}, and formally verified delegation protocols \citep{chen2026aith}. The gap between these fragments and a unified authorization propagation architecture is the central open challenge.

The central claim is that identity governance must be treated as infrastructure --- evaluated continuously, enforced at every interaction boundary, and designed into the system from the start. Companion work on machine-speed accountability \citep{tallam2026machine}, fail-and-report \citep{tallam2026failandreport}, execution envelopes \citep{tallam2026executionenvelope}, and authorization-limited evidence evaluation \citep{tallam2026partialbench} sketches adjacent pieces of that infrastructure. The alternative --- treating authorization as a policy layer applied after the agent architecture is built --- produces systems that are structurally prone to unauthorized data exposure through normal operation, independent of any adversarial action.

\section*{Acknowledgments}

This paper benefited from informal discussions with colleagues working on agentic AI security, model evaluation, and enterprise authorization systems. The SQL injection analogy in Section 8.1 was crystallized in a particularly heated conversation about whether prompt injection is the only novel security problem in agentic AI. The answer, as this paper argues, is: it is the only novel \emph{attack vector}. It is not the only novel \emph{problem}.

\bibliographystyle{plainnat}
\bibliography{references}

@unpublished{tallam2026machine,
  title = {Machine-Speed Accountability: A Constraint-Aware Reference Model for Oversight-Heavy AI},
  author = {Tallam, Krti},
  year = {2026},
  note = {Working draft}
}

@unpublished{tallam2026failandreport,
  title = {Fail-and-Report: A Missing Authorization Primitive for Agentic AI Systems},
  author = {Tallam, Krti},
  year = {2026},
  note = {Working draft}
}

@unpublished{tallam2026partialbench,
  title = {Partial Evidence Bench: Benchmarking Authorization-Limited Evidence in Agentic Systems},
  author = {Tallam, Krti},
  year = {2026},
  note = {Working draft}
}

@unpublished{tallam2026executionenvelope,
  title = {Execution Envelopes: A Shared Admission Contract for Backend AI Execution Requests},
  author = {Tallam, Krti},
  year = {2026},
  note = {Working draft}
}

@unpublished{tallam2026fromcantowould,
  title = {From Can to Would: Identity-Conditioned Authorization for Delegated Agentic Action},
  author = {Tallam, Krti},
  year = {2026},
  note = {Working draft}
}

@unpublished{tallam2026layeredmutability,
  title = {Layered Mutability: Identity Drift and Governance in Self-Modifying AI Agents},
  author = {Tallam, Krti},
  year = {2026},
  note = {Working draft}
}

@misc{deepmind2026agenttraps,
  title = {{AI Agent Traps}: A Taxonomy of Attacks on {AI} Agents},
  author = {{Google DeepMind}},
  year = {2026},
  howpublished = {Technical report},
  key = {DeepMind}
}

@inproceedings{pang2019zanzibar,
  title = {Zanzibar: Google's Consistent, Global Authorization System},
  author = {Pang, Ruoming and Caceres, Ramon and Burrows, Mike and Chen, Zhifeng and Dave, Pratik and Germer, Nathan and Golynski, Alexander and Graney, Kevin and Kang, Nina and Kissner, Lea and others},
  booktitle = {USENIX Annual Technical Conference (ATC)},
  year = {2019}
}

@techreport{nist2023airmf,
  title = {Artificial Intelligence Risk Management Framework ({AI RMF} 1.0)},
  author = {{National Institute of Standards and Technology}},
  year = {2023},
  institution = {NIST},
  number = {AI 100-1},
  key = {NIST}
}

@techreport{iso2023iec42001,
  title = {{ISO/IEC} 42001:2023 Information Technology --- Artificial Intelligence --- Management System},
  author = {{International Organization for Standardization}},
  year = {2023},
  institution = {ISO/IEC},
  key = {ISO}
}

@misc{gruskovnjak2023indirect,
  title = {Indirect Prompt Injection via Web Search},
  author = {Gruskovnjak, Simon},
  year = {2023},
  howpublished = {Technical writeup}
}

@inproceedings{greshake2023compromising,
  title = {Not What You've Signed Up For: Compromising Real-World {LLM}-Integrated Applications with Indirect Prompt Injection},
  author = {Greshake, Kai and Abdelnabi, Sahar and Mishra, Shailesh and Endres, Christoph and Holz, Thorsten and Fritz, Mario},
  booktitle = {Proc. 16th ACM Workshop on Artificial Intelligence and Security (AISec)},
  year = {2023}
}

@misc{prakash2026aip,
  title = {{AIP}: Invocation-Bound Capability Tokens with {Biscuit} and {Datalog}},
  author = {Prakash, Sunil},
  year = {2026},
  eprint = {2603.24775},
  archivePrefix = {arXiv}
}

@misc{sharma2026pauth,
  title = {{PAuth}: Precise Task-Scoped Authorization for {AI} Agents},
  author = {Sharma, Aditya and Jiang, Ruoyu and Lin, Zheng and Chen, Linda},
  year = {2026},
  eprint = {2603.17170},
  archivePrefix = {arXiv}
}

@misc{parakhin2026bureaucracy,
  title = {The Bureaucracy of Speed: Capability Coherence Systems for Agent Authorization},
  author = {Parakhin, Vladyslav},
  year = {2026},
  eprint = {2603.09875},
  archivePrefix = {arXiv}
}

@misc{palumbo2026pcas,
  title = {{PCAS}: Policy Compiler for Agentic Systems},
  author = {Palumbo, Alessandro and Jha, Somesh and others},
  year = {2026},
  eprint = {2602.16708},
  archivePrefix = {arXiv}
}

@misc{chen2026aith,
  title = {{AITH}: Post-Quantum Continuous Delegation for {AI} Agents},
  author = {Chen, Zhaoliang},
  year = {2026},
  eprint = {2604.07695},
  archivePrefix = {arXiv}
}

@misc{garby2026llmbda,
  title = {{LLMbda} Calculus: Dynamic Information-Flow Control for Agentic Programming},
  author = {Garby, Sebastian and Gordon, Andrew D. and Sands, David},
  year = {2026},
  eprint = {2602.20064},
  archivePrefix = {arXiv}
}

@misc{song2026formalizing,
  title = {Formalizing {LLM} Agent Security},
  author = {Song, Dawn and others},
  year = {2026},
  eprint = {2603.19469},
  archivePrefix = {arXiv},
  note = {UC Berkeley / ETH Zurich}
}

@misc{openid2025agentic,
  title = {Agentic Identity Whitepaper},
  author = {{OpenID Foundation}},
  year = {2025},
  eprint = {2510.25819},
  archivePrefix = {arXiv},
  key = {OpenID}
}

@misc{benameur2025oidca,
  title = {{OIDC-A}: OpenID Connect for Agents 1.0},
  author = {Benameur, Azzedine and others},
  year = {2025},
  eprint = {2509.25974},
  archivePrefix = {arXiv}
}

@misc{ahad2026semantic,
  title = {Semantic Intent Fragmentation: Aggregation Attacks on Agentic {AI} Systems},
  author = {{Anonymous}},
  year = {2026},
  eprint = {2604.08608},
  archivePrefix = {arXiv},
  note = {Accepted AAAI 2026 Summer Symposium},
  key = {SIF}
}

@misc{li2026agentic,
  title = {Agentic Deanonymization},
  author = {Li, Tianshi},
  year = {2026},
  eprint = {2601.05918},
  archivePrefix = {arXiv}
}

@misc{maiti2026healthcare,
  title = {Healthcare Zero Trust: Production Deployment of Autonomous {AI} Agents},
  author = {Maiti, Saikat},
  year = {2026},
  eprint = {2603.17419},
  archivePrefix = {arXiv}
}

@misc{mcpshield2026,
  title = {{MCPSHIELD}: Threat Analysis of {MCP} Tool Ecosystem},
  author = {{Anonymous}},
  year = {2026},
  eprint = {2604.05969},
  archivePrefix = {arXiv},
  key = {MCPSHIELD}
}

@misc{seagent2026,
  title = {{SEAgent}: Mandatory Access Control for Multi-Agent {LLM} Systems},
  author = {{Anonymous}},
  year = {2026},
  eprint = {2601.11893},
  archivePrefix = {arXiv},
  key = {SEAgent}
}

@misc{debenedetti2026roleconfusion,
  title = {Prompt Injection as Role Confusion},
  author = {Debenedetti, Giulio and others},
  year = {2026},
  eprint = {2603.12277},
  archivePrefix = {arXiv},
  note = {MIT}
}

@misc{anthropic2026claudecode,
  title = {{Claude Code} Permission System Evaluation},
  author = {{Anonymous}},
  year = {2026},
  eprint = {2604.04978},
  archivePrefix = {arXiv},
  key = {ClaudeCode}
}

@misc{bhattarai2026deterministic,
  title = {Deterministic Architectural Boundaries for Agentic {AI}},
  author = {Bhattarai, Prabin and Vu, Tuan},
  year = {2026},
  eprint = {2602.09947},
  archivePrefix = {arXiv}
}

@misc{errico2026aarm,
  title = {{AARM}: Action Classification with Tamper-Evident Receipts},
  author = {{Anonymous}},
  year = {2026},
  eprint = {2602.09433},
  archivePrefix = {arXiv},
  key = {AARM}
}

@misc{ruohonen2025sok,
  title = {{SoK}: Attack Surface of Agentic {AI}},
  author = {{Anonymous}},
  year = {2026},
  eprint = {2603.22928},
  archivePrefix = {arXiv},
  key = {SoK-Agentic}
}

@techreport{bell1973secure,
  title = {Secure Computer Systems: Mathematical Foundations},
  author = {Bell, David Elliott and LaPadula, Leonard J.},
  year = {1973},
  institution = {MITRE Corporation},
  number = {MTR-2547}
}

@techreport{biba1977integrity,
  title = {Integrity Considerations for Secure Computer Systems},
  author = {Biba, Kenneth J.},
  year = {1977},
  institution = {MITRE Corporation},
  number = {MTR-3153}
}

@inproceedings{clark1987comparison,
  title = {A Comparison of Commercial and Military Computer Security Policies},
  author = {Clark, David D. and Wilson, David R.},
  booktitle = {Proc. IEEE Symposium on Security and Privacy},
  year = {1987}
}

@article{sandhu1996role,
  title = {Role-Based Access Control Models},
  author = {Sandhu, Ravi S. and Coyne, Edward J. and Feinstein, Hal L. and Youman, Charles E.},
  journal = {IEEE Computer},
  volume = {29},
  number = {2},
  pages = {38--47},
  year = {1996}
}

@techreport{xacml2013,
  title = {{eXtensible} Access Control Markup Language ({XACML}) Version 3.0},
  author = {{OASIS}},
  year = {2013},
  institution = {OASIS Standard},
  key = {XACML}
}

\end{document}